\documentclass[11pt,a4paper]{article}
\usepackage[hyperref]{emnlp2020}
\usepackage{times}
\usepackage{latexsym}
\usepackage{graphicx}
\usepackage{multirow}
\usepackage{booktabs}

\usepackage{subcaption}
\usepackage{xcolor}
\usepackage{tikz}
\usetikzlibrary{positioning}
\usetikzlibrary{arrows.meta}

\usepackage{microtype}

\newcommand{\mycbox}[1]{\tikz{\path[draw=#1,fill=#1] (0,0) rectangle (0.2cm,0.2cm);}}

\newcommand\blfootnote[1]{%
  \begingroup
  \renewcommand\thefootnote{}\footnote{#1}%
  \addtocounter{footnote}{-1}%
  \endgroup
}

\title{TaxiNLI: Taking a Ride up the NLU Hill}

\author{
    Pratik Joshi$^{2\dagger*}$, 
    Somak Aditya$^{1\dagger}$, 
    Aalok Sathe$^{3\dagger*}$, 
    \and Monojit Choudhury$^{1}$ \\
  $^{1}$Microsoft Research India, 9 Lavelle Road, Vigyan, Bengaluru, India\\
  $^{2}$Google Research, Carina East Tower Bagmane Constellation Business Park, Bengaluru, India\\
  $^{3}$University of Richmond, 410 Westhampton Way, Richmond, VA, USA \\
  \resizebox{\textwidth}{!}
        {
            pratikmjoshi123@gmail.com,
            aalok.sathe@richmond.edu,
            \{t-soadit,monojitc\}@microsoft.com
        }
}

\date{}

\begin{document}
\aclfinalcopy
\maketitle
\blfootnote{$\dagger$ denotes equal contribution. $^*$Work was done while Authors were at Microsoft Research India.}

\begin{abstract}
  Pre-trained Transformer-based neural architectures have consistently achieved state-of-the-art performance in the Natural Language Inference (NLI) task. 
  Since NLI examples encompass a variety of linguistic, logical, and reasoning phenomena, it remains unclear as to which specific concepts are learnt by the trained systems and where they can achieve strong generalization. To investigate this question, 
  we propose a taxonomic hierarchy of categories that are relevant for the NLI task. We introduce  \textsc{TaxiNLI}, a new dataset, that has 10k examples from the MNLI dataset~\cite{bowman2018-mnli} with these taxonomic labels. 
  Through various experiments on \textsc{TaxiNLI}, we observe that whereas for certain taxonomic categories SOTA neural models have achieved near perfect accuracies---a large jump over the previous models---some categories still remain difficult. Our work adds to the growing body of literature that shows the gaps in the current NLI systems and datasets through a systematic presentation and analysis of reasoning categories. 
\end{abstract}
\section{Introduction}

The Natural Language Inference (NLI) task tests whether a hypothesis (H) in text contradicts with, is entailed by, or is neutral with respect to a given premise (P) text. 
This 3-way classification task, popularized by \citet{bowman2015large}, which was in turn inspired by \citet{dagan2005pascal},  now serves as a benchmark for \textit{evaluation} of natural language \textit{understanding} (NLU) capability of models; for example, NLI datasets~\cite{bowman2015large,bowman2018-mnli} are included in all NLU benchmarks  such as GLUE and SuperGLUE \cite{wang2018glue}. These corpora, in turn, have been successfully used to train models such as BERT \cite{devlin2019bert} to achieve state-of-the-art (SOTA) performance in these tasks. Despite the wide adoption of NLI datasets, a growing concern in the community has been the lack of clarity as to \textit{which linguistic or reasoning concepts these trained NLI systems are truly able to learn and generalize} (see, for example~\cite{linzen-2020-accelerate} and \cite{bender-koller-2020-climbing}, for a discussion). Over the years, as models have shown steady performance increases in NLI tasks, many authors \cite{nie2019adversarial,kaushik2019learning} demonstrate steep drops in performance when these models are tested against adversarially (or counterfactually) created examples by non-experts.  \citet{Richardson2019ProbingNL} use templated examples to show trained NLI systems fail to capture essential logical (negation, boolean, quantifier) and semantic (monotonicity) phenomena.

    Herein lie the central questions of our work: 1) what is the distribution of various categories of {\em reasoning tasks} in the NLI datasets? 2) which categories of tasks are rarely captured by current NLI datasets (owing to the nature of the task and the non-expert annotators)? 3) which categories are well-understood by the SOTA models? and 4) are there categories where Transformer-based architectures are consistently deficient? 
    
    In order to answer these questions, we first discuss why performance-specific error analysis categories \cite{wang2018glue, nie2019adversarial}, and stress testing categories \cite{naik-etal-2018-stress} are inadequate. We then propose a taxonomy of the various reasoning tasks that are commonly covered by the current NLI datasets (Sec~\ref{sec:taxonomy}).  Next, we annotate 10,071 P-H pairs from the MNLI dataset~\cite{bowman2018-mnli} with the lowest level taxonomic categories, 18 in total (Sec~\ref{sec:taxinli}). Then we conduct various experiments and careful error analysis of the SOTA models---BERT and RoBERTa, as well as other baselines, such as Bag-of-words Na\"ive Bayes and ESIM, on their performance across these categories (Sec~\ref{sec:eval}). Our analyses indicate that while these models perform well on some categories such as linguistic reasoning, the performance on many other categories, such as those that require world knowledge or temporal reasoning, are quite poor. We also look into the embeddings of the P-H pairs to understand which of these categorical distinctions are captured well in the learnt representations, and which get conflated (Sec~\ref{sec:discussion}). Inline with our previous finding, we observe strong correlation between the level of clustering within the representation of the examples from a category, and the performance of the models for that particular category.

\section{A New Taxonomy for NLI}
\label{sec:taxonomy}
\subsection{Necessity for a New Taxonomy}
According to \citet{wittgenstein-1922}, ``Language disguises the thought'', and human beings try to gauge such thought from colloquial language using  ``complex silent adjustments''. The journey from lexicon and syntax of ``language'' to the aspects of semantics and pragmatics can be thought of as a journey that 
portrays important milestones that an ideal NLU system should achieve. Irrespective of the order of such milestones\footnote{``For an infant, a foreigner, or an instant-message addict, context is more important than syntax'' \cite{sowa2010role}.}, we believe that NLU (and NLI) systems should be tested and analyzed with respect to fundamental linguistic and logical phenomena. Recently, different types of phenomena have been tested through 1) creating new datasets, 2) probing tasks, and 3) error-analysis categorizations. Researchers have created new datasets by recasting various NLU tasks to a large NLI dataset \cite{poliak-etal-2018-collecting}, eliciting counter-factual examples from non-experts by considering different lexical and reasoning factors \cite{kaushik2019learning}, and adversarial example \cite{nie2019adversarial} elicitation by letting non-experts come up with examples through interacting with SOTA systems. However, these datasets do not expose the linguistic aspects where the current systems have difficulty. Using the probing task methodology, researchers \cite{jawahar-etal-2019-bert,goldberg2019assessing} observed that BERT captures syntactic structure, along with some semantics such as NER, and semantic role labels \cite{tenney2019you}. However, BERT's ability to reason is questioned by the observed performance degradation in MNLI \cite{mccoy-etal-2019-right}. \citeauthor{linzen-2020-accelerate} (\citeyear{linzen-2020-accelerate}) also called for a pretraining-agnostic evaluation setup, where the setup is not limited to pre-trained language models. Our taxonomic categorization is meant to serve as a set of necessary inferencing capabilities that one would expect a competing NLI system to possess; thereby promoting more probing tasks along unexplored categories. 
\begin{figure}
	\centering
    \includegraphics[width=0.48\textwidth]{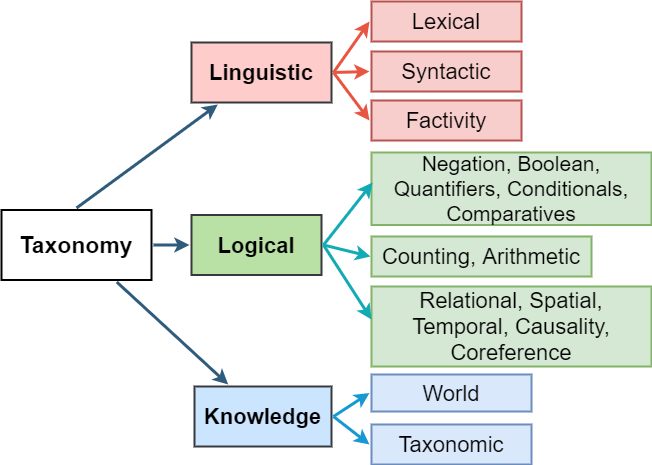} 
    \caption{Taxonomic Categorization of the NLI task. 
    }
    \label{tax1}
\end{figure}

  Existing categorization efforts have centred around informing feature creation in the pre-Transformer era, and model-specific error analysis in more recent times. Previously, \cite{lobue2011types} enumerated the type of commonsense knowledge required for NLI. 
  Among recent error analysis efforts, the GLUE diagnostic dataset \cite{wang2018glue}, inference types for Adversarial NLI \cite{nie2019adversarial}, the new CheckList \cite{ribeiro-etal-2020-beyond} system and the Stress Tests \cite{naik-etal-2018-stress} are mentionworthy.  
  As we attempted to group the categorizations in \citet{nie2019adversarial} and \citet{wang2018glue} into four high-level categories (lexical, syntactic, semantic, and pragmatic)\footnote{Table provided in Appendix}, we observe that there is a lack of consensus, non-uniformity and repetitiveness of these categories.  For example, the \textit{Tricky} label in \citet{nie2019adversarial} groups examples that involve ``wordplay, linguistic strategies such as syntactic transformations, or inferring writer intentions from contexts''; thereby spanning aspects of syntax and pragmatics. Similarly, \textit{Reference and Names} requires both reasoning and knowledge. The GLUE diagnostic categories \cite{wang2018glue} does not include interesting reasoning categories such as \textit{temporal}, and \textit{spatial}. 
  The stress types proposed by \citet{naik-etal-2018-stress} are specific to mostly lexical and some semantic corner cases. This is expected, as these categorizations are analysis-oriented and often dependent on the performance of a set of models in question. Here, we propose a taxonomic categorization that delineates a set of necessary uniform inferencing capabilities for the NLI task.

  \begin{table*}[!htpb]
  \resizebox{\textwidth}{!}{  
\begin{tabular}{p{2.5cm}lp{2.5cm}l}
\toprule
            \textbf{Taxonomic Category} & \textbf{MNLI Examples }                                                                                                                                         &    \textbf{Taxonomic Category}           & \textbf{MNLI Examples }                                                                                                                                                   \\ \midrule
{\texttt{Lexical}}      & \begin{tabular}[c]{@{}l@{}}P: so it's stayed cold for the \underline{entire} week \\ H: It has been cold for the \underline{whole} week.\end{tabular}                                                                                                                                                                                             & {\texttt{Relational}}  & \begin{tabular}[c]{@{}l@{}}P: Actually, \underline{my sister} wrote a story on it. \\ H: \underline{My sibling} created a story about it.\end{tabular}                                                                                                                                                          \\ \hline
{\texttt{Syntactic}}    & \begin{tabular}[c]{@{}l@{}}P: Those in Egypt, Libya, Iraq, and Yemen were \\ eventually overthrown by secular nationalist revolutionaries.\\ H: Secular nationalist revolutionaries eventually \\ overthrew them in Egypt and Libya.\end{tabular}                                                                                                                                   & {\texttt{Spatial}}     & \begin{tabular}[c]{@{}l@{}}P: At the eastern end of Back Lane and turning right,\\ Nicholas Street becomes Patrick Street, and in St. Patrick's\\ Close is St. Patrick's Cathedral .\\ H: Nicholas Street becomes Patrick Street after\\ turning left at the eastern end of Back Lane.\end{tabular}                                               \\ \hline
{\texttt{Factivity}}    &                                \begin{tabular}[c]{@{}l@{}}P: The best place to view the spring azaleas is at the Azalea Festival\\ in the last week of April at Tokyo's Nezu shrine.\\ H:There is an Azalea Festival at the Nezu Shrine.\end{tabular}                                                                                                                                                                                                                                                                                                                                   & {\texttt{Temporal}}    & \begin{tabular}[c]{@{}l@{}}P: See you \underline{Aug. 12, or soon thereafter}, we hope.\\ H: The person told not to come \underline{until December}.\end{tabular}                                                                                                                                               \\ \hline
{\texttt{Negation}}     & \begin{tabular}[c]{@{}l@{}}P: They \underline{post} loads of newspaper articles--Yahoo!\\ H:  Yahoo does \underline{not post} any articles from newspapers.\end{tabular}                                                                                                                                                                          & {\texttt{Causal}}      & \begin{tabular}[c]{@{}l@{}}P: Acroseon the mountainside is another terrace on which \\ imperial courtiers and dignitaries would sit while \\ enjoying dance performances and music recitals on the \\ \underline{hondo's broad terrace}.\\ H: There is a \underline{terrace} where musicians play.\end{tabular} \\ \hline

{\texttt{Boolean}}      & \begin{tabular}[c]{@{}l@{}}P:According to contemporaneous notes, at 9:55 the Vice President \\ was still on the phone with the President advising that three planes \\were missing \underline{and} one had hit the Pentagon.\\ H:  The President called the Vice President to tell him the plane \\hit the Pentagon.\end{tabular} & \texttt{Coreference}       & \begin{tabular}[c]{@{}l@{}}P: A dozen minor wounds crossed his forearms and body. \\ H: The grenade explosion left him with a lot of wounds.\end{tabular}                                                                                                                                                                                         \\ \hline
{\texttt{Quantifier}}   & \begin{tabular}[c]{@{}l@{}}P: \underline{Some} travelers add Molokai and Lanai to their itineraries.\\ H:  \underline{No one} decides to go to Molokai and Lanai.\end{tabular}                                                                                                                                                                     & \texttt{World}       & \begin{tabular}[c]{@{}l@{}}P: In this respect, bringing Steve Jobs back to save Apple\\ is like bringing Gen.\\ H: Steve Jobs unretired in 2002.\end{tabular}                                                                                                                                                                                                                                             \\ \hline
{\texttt{Conditional}} & \begin{tabular}[c]{@{}l@{}}P: If the revenue is transferred to the General Fund, it is recognized \\as nonexchange revenue in the Government-wide consolidated \\ financial statements.\\ H: Revenue from the General Fund is not considered in financial statements\end{tabular}                                                                                                                                                                                                   & \texttt{Taxonomic} & \begin{tabular}[c]{@{}l@{}}P: Benson's action picture in Lucia in London \\(Chapter 8)- Georgie stepped on a beautiful \underline{pansy}. \\ H: Georgie crushed a beautiful \underline{flower} in Chapter 8 \\of Lucia in London.\end{tabular}                                                                              \\ \hline
\texttt{Comparative} & \begin{tabular}[c]{@{}l@{}}P: Load time is divided into elemental and coverage related load time.\\ H: The coverage related load time \underline{is longer than} elemental.\end{tabular}                     &             &                                                           \\ \bottomrule
\end{tabular}
}
\caption{For each category, we provide an example from the MNLI dataset. For a full set of synthetic examples and definitions, please look at appendix.}
\label{tabexamples}
\end{table*}

\subsection{Taxonomic Categories: Definitions and Examples}
In Figure \ref{tax1}, we present our taxonomic categorization. Our categorization is based on the following principles.  \underline{First}, we take a model-agnostic approach, where we work from the first principles to arrive at a set of basic inferencing processes that are required in NLI task. These include an unrestricted variety of linguistic and logical phenomena, and may require knowledge beyond text, thus providing us with the higher-level categories: \textit{linguistic}, \textit{logical} and \textit{knowledge}-based. \underline{Second}, we retain categories that are non-overlapping and sufficiently represented in NLI datasets. For example, for sub-categories under {\it linguistic}, we prune \textit{semantics} because necessary aspects are covered by \texttt{logical} and \texttt{knowledge}-based categories. We omit specific aspects of \textit{pragmatics} such as implicatures and pre-suppositions, as they are rarely observed in NLI datasets \cite{jeretic-etal-2020-natural}. \underline{Thirdly}, we aim to list a set of necessary sub-categories. For example, for logical deduction sub-categories, we take inspiration from \citet{davis2015commonsense}, who list the commonsense reasoning categories where systems have seen success.   \underline{Lastly}, since we aim to employ non-experts for collecting annotations, we decide to restrict further sub-division wherever the definitions get complicated, or pre-suppose certain expertise; for example the \texttt{lexical} category is not sub-divided further (as followed in \citet{wang2018glue}). Thus, we take a pragmatic approach that is theory neutral and does not warrant coverage of all reasoning tasks, though we do believe that the taxonomy is sufficiently deep and generic that allows systematic and meaningful analysis of NLI models with respect to their reasoning capabilities.

Next we define the categories. For a full set of examples, please see Table \ref{tabexamples}. 


\underline{High-Level Categories}: The \textbf{\texttt{Linguistic}} category represents NLI examples where the inference process to determine the entailment are internal to the provided text. We classify examples as \textbf{\texttt{Logical}} when the inference process may involve processes external to text, such as mapping words to percepts and reason with them \cite{sowa2010role}. \textbf{\texttt{Knowledge}}-based category represents examples where some form of external, domain or commonly assumed knowledge is required for inferencing. 

\textbf{\texttt{Linguistic}} category is further sub-divided into \texttt{lexical}, \texttt{syntactic}, and \texttt{factivity}.\\\noindent
1. \underline{\textbf{Lexical}}: This category captures P-H pairs where the text is almost the same apart from removal, addition or substitution of some lexical items. \textbf{Example}: P: Anakin was kind. H: Anakin was cruel.\\\noindent
2. \underline{\textbf{Syntactic}}:  \texttt{Syntactic} deals with examples where syntactic variations or paraphrases are essential to detecting entailment. \textbf{Example}: P: Anakin was an excellent pilot. H: The piloting skills of Anakin were excellent.\\\noindent
3. \underline{\textbf{Factivity}}:  Here the hypothesis contains an assumed fact from the premise, mostly an assumption about the existence of an entity or the occurrence of an action (inspired from \citet{wang2018glue}). \textbf{Example}: P: Padme recognized that Anakin was intelligent. H: Anakin was intelligent.

Based on commonalities, \textbf{\texttt{Logical}} categories are grouped under ``Connectives''. ``Mathematical'' and ``Deduction''. \\\noindent
1. \underline{\textbf{Connectives (Negation, Boolean, Quantifiers}}, \underline{\textbf{Conditionals, Comparatives)}}: We group the logical categories \texttt{negation, boolean, quantifier, conditional} and \texttt{comparative} \cite{salvatore2019logical} under the ``Connectives'' label. \texttt{Negation} applies when P negates one (or more) of the facts in H. We apply the category \texttt{boolean} when P is a set of statements connected by \textit{or, and} and H talks about one of the statements. \texttt{Quantifier} is applied when P or H requires understanding of words denoting existential or universal quantifiers. Similarly, \texttt{conditional} is applicable where P or H has conditional statements. If P (or H) compares entities via comparative phrases, then we label it as \texttt{comparative}. \textbf{Examples}: (\texttt{boolean} and \texttt{negation}) P: Jar Jar, R2D2 and Padme only visited Anakin’s house. H: Jar Jar Binks didn’t visit Anakin’s shop.\\\noindent
2. \underline{\textbf{Mathematical (Counting, Arithmetic)}}: This group of categories is concerned with examples that require mathematical reasoning. For brevity, we concentrate on examples that require counting and simple arithmetic operations. However, we observed exceedingly low number of examples in this category group from our pilot study on SNLI and MNLI, and hence we remove these from our final annotations.\\\noindent
3. \underline{\textbf{Deductions (Relational, Spatial, Temporal}}, \underline{\textbf{Causal, Coreference)}}: Motivated by predicate logic, success of qualitative representation and reasoning in dealing with temporal and spatial reasoning \cite{gabelaia2005combining}, and causality \cite{pearl2009causality}, we list \texttt{relational}, \texttt{temporal, spatial} and \texttt{causal} under ``Deductions''. The \texttt{relational} reasoning stands for the requirement to perform deductive reasoning using relations present in text. \texttt{Spatial} (and \texttt{temporal}) denotes reasoning using spatial (and temporal) properties of objects represented in text.  We also consider language-inspired reasoning categories such as co-reference resolution, which is known to often require event-understanding \cite{ng2017machine} beyond superficial cues. \textbf{Example}: (\texttt{relational}) P: The lamp was working properly. H: The lightbulb from the lamp was not functioning.

Lastly, we define two sub-categories under \textbf{\texttt{Knowledge}}, namely \texttt{world} and \texttt{taxonomic}.\\\noindent
1. \underline{\textbf{World}}: Examples that require knowledge about named entities, knowledge about historical, current events; and domain-specific knowledge. \textbf{Example}: (\texttt{world}) P: Michelle Obama stayed in the White House during 2009-17. H: Michelle was living in the White House legally during 2009. \\\noindent
2. \underline{\textbf{Taxonomic}}: Examples that require taxonomies and hierarchies. For example, \textit{IsA, hasA, hasProperty} relations. \textbf{Example}: (\texttt{taxonomic}) P: Norman hated all musical instruments. H: Norman loves the piano.

Note that presence of a certain lexical trigger for a category (such as negation) does not warrant the labeling with the category, unless understanding of that concept is invoked in the deduction process.

\section{TaxiNLI: Dataset Details}
\label{sec:taxinli}

We present TaxiNLI, a dataset collected based on the principles and categorizations of the aforementioned taxonomy. We curate a subset of examples from MultiNLI \cite{bowman2018-mnli} by sampling uniformly based on the entailment label and the domain. We then annotate this dataset with fine-grained category labels.  

\subsection{Annotation Process}
 

\paragraph{Task Design} For large-scale data collection, our aim was to propose an annotation methodology that is relatively flexible in terms of annotator qualifications, and yet results in high quality annotations. To employ non-expert annotators, we designed a simplified guideline (questionnaire/interface) for the task, that does not pre-suppose expertise in language or logic. 
As an overhead, the guideline requires a few rounds of one-on-one training of the annotators. Because it is expensive to perform such rounds of training in most crowdsourcing platforms, we hire and individually train a few chosen annotators.
Upon conducting the previously-discussed pilot study and using the given feedback, we created a hierarchical questionnaire which first asked the annotator to do the NLI inference 
on the P-H pair, and then asked targeted questions to get the desired category annotations for the datapoints. The questionnaire is shared in the Appendix.

For the MNLI datapoints with `neutral' gold labels, we realized, through observation and annotator feedback, that annotating the categories were difficult, as sometimes the hypotheses could not be connected well back to their premise. Hence, we created 2 questionnaires, one for the `entailment/contradiction' examples, and one for `neutral' examples. For the entailment/contradiction examples, We collected binary annotations for each of the 15 categories in our NLI taxonomy, for datapoints in MNLI which had `entailment' or `contradiction' as gold labels.  
To resolve this, for the `neutral' examples we specifically 
asked them whether the premise and hypothesis were discussing 1) the same general topic (politics, geology, etc.), and if so, 2) had the same subject and/or object of discussion (Obama, Taj Mahal, etc.). If the response to 2) was `yes', then they were asked to provide the category annotations as previously defined. 

\paragraph{Annotator Training/Testing} We first tested our two annotators by asking them to do inference on a set of randomly selected premise-hypothesis pairs from MultiNLI. This was to familiarize them with the inference task. After giving the category annotation task, we also continuously tested and trained the two annotators. After a set of datapoints were annotated, we reviewed and went through clarification and feedback sessions with the annotators to ensure they understood the task, and the categories, and what improvements were required. More details are provided in the Appendix. 
 
 
 \subsubsection{Annotation Metrics}
\begin{figure}
	\centering
    \includegraphics[width=0.5\textwidth]{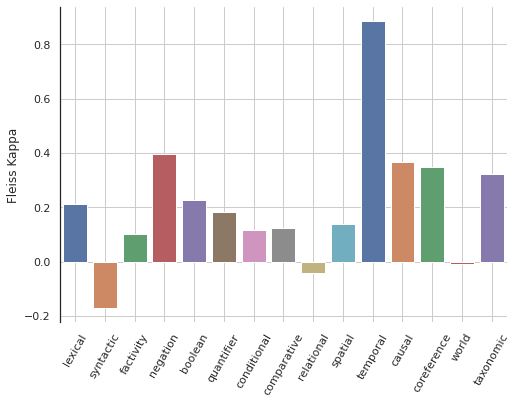}
    \caption{Inter-annotator Agreement (IAA) values between the two annotators, plotted category-wise.}
    \label{fig:iaa_categories}
\end{figure}

 Here, we assess the individual annotator performance and inter-annotator agreement. Since, automated metrics for individual complex category annotations are hard to define, we use an indicative metric that matches the annotated inference label with the gold label, i.e., their \textbf{inference accuracy}.
 We also calculated inter-annotator agreement between the two annotators for an overlapping subset of 600 examples. For agreement, we use the Fleiss' Kappa ($\kappa$) \cite{fleiss}. We also compute another simple statistic, namely the `IOU' (Intersection-Over-Union) of categories per datapoint, defined as:
 $IOU = \frac{1}{N}\sum_{i=1}^{N}{\frac{|{C_{A1}^{i}}\cap{C_{A2}^{i}}|}{|{C_{A1}^{i}}\cup{C_{A2}^{i}}|}},$ where $C_{Ai}^{j}$ are the category annotations for Annotator $i$ for datapoint $j$, averaged over total datapoints $N$. 
 Looking at the category-wise Fleiss' $\kappa$ values in Fig. \ref{fig:iaa_categories}, we observe that there are promising levels of agreement in most of the categories except \texttt{syntactic}, \texttt{relational}, and \texttt{world}. We observe the average inference accuracy (86.7\%) is high despite known issues in MNLI example ambiguity. Similarly, both the average Fleiss' $\kappa$ (\textbf{0.226}) and the IOU metric (\textbf{0.241}) suggest an overall reasonable inter-annotator agreement.
 
\subsection{Dataset Statistics}

Each datapoint in TaxiNLI\footnote{The dataset is available for download from \url{https://github.com/microsoft/TaxiNLI}.} consists of a premise-hypothesis pair, the entailment label, and binary annotations for 18 features. 15 features correspond to the 15 categories discussed in the taxonomy, and 3 additional features for the `neutral' gold label datapoints based on same general topic, same subject, and same object. The statistics are listed in Tab. \ref{tab:taxinlistats2}.

\begin{table}[!h]
\begin{center}
\resizebox{\columnwidth}{!}{
\begin{tabular}{ll}
      \toprule
      Total datapoints& 10,071 \\
      \midrule
      \multirow{2}{*}{
        Datapoints overlapping with MNLI
      } 
        & 2343
        (train) \\
        &7728 (dev)  \\
      \hline
      Avg. datapoints per domain& $1007.1$ \\
      \hline
      Datapoints per NLI label & 3375 (C), 3201 (N), 3495 (E) \\
      \hline
      Avg. categories per datapoint & $1.6$ \\
      \hline
      \multirow{3}{*}{
        Neutral example characteristics
      } 
        & 3087 (Same general topic),\\ 
        & 2843 (Same object),\\
        & 877 (Same subject) \\
      \bottomrule
\end{tabular}}
\caption{TaxiNLI Statistics}
\label{tab:taxinlistats2}
\end{center}
\end{table}

\begin{figure}
	\centering
    \includegraphics[width=0.48\textwidth]{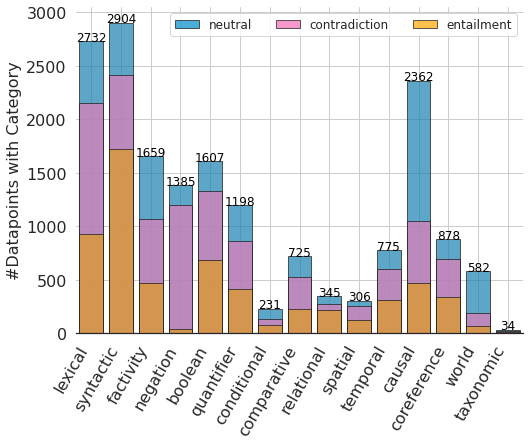}
    \caption{The number of datapoints annotated with each category, split by the gold label of the datapoints. For \texttt{taxonomic}, the gold label split is 9(E)/7(C)/3(N).}
    \label{fig:categories_freq}
\end{figure}

\paragraph{Categorical Observations} From our annotations, we observe that inferencing each MNLI example requires about 2 categories. Fig.~\ref{fig:categories_freq} shows the distribution of categories in the TaxiNLI dataset. We see that a large number of P-H pairs in MNLI require \texttt{lexical} and \texttt{syntactic} knowledge to make an inference; whereas the challenges of \texttt{relational}, \texttt{spatial}, and \texttt{taxonomic} for inference are not adequately represented. There is a large proportion of examples in the \texttt{syntactic} category which have the `entailment' label, and a large proportion of \texttt{negation} examples have the `contradiction' label. Additionally, many `neutral' examples were classified as requiring \texttt{causal} knowledge. The feedback session with annotators revealed that there were many `neutral' examples where the hypothesis was essentially an unverifiable intent or detail of a certain action mentioned in the premise. An example is ``\textbf{P: }\textit{Another influence was that patrician politician Franklin Roosevelt, who was, like John D. Rockefeller, the focus of Nelson's relentless sycophancy and black-belt bureaucratic infighting.} \textbf{H: }\textit{Nelson targeted Roosevelt in order to gain political favor.}".

\begin{figure}[htb!]
	\centering
    \includegraphics[width=0.5\textwidth]{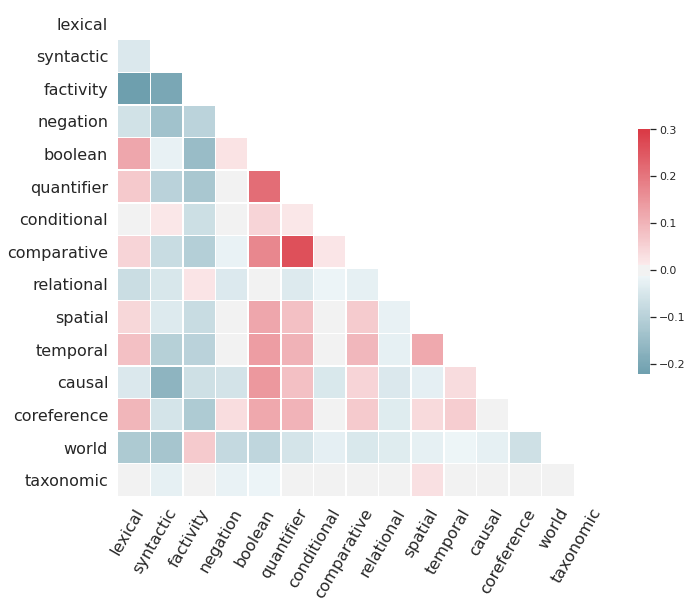}
    \caption{The correlation matrix between the taxonomic categories. }
    \label{categorical_correl}
\end{figure}

\paragraph{Categorical Correlations} Fig.~\ref{categorical_correl} shows correlations among categories in our dataset. We observe that most categories show weak correlation in the MNLI dataset, hinting at a possible independence of categories with respect to each other. Relatively stronger positive correlations are seen between \texttt{boolean}-\texttt{quantifier}, and \texttt{boolean}-\texttt{comparative} categories. We specifically looked at the genre-wise split of datapoints containing \texttt{boolean}-\texttt{quantifier} and saw that nearly 25\% of them came from the `telephone' genre of MNLI. An example is ``\textbf{P: }\textit{have that well and it doesn't seem like very many people uh are really i mean there's a lot of people that are on death row but there's not very many people that actually um do get killed} \textbf{H: }\textit{Most people on death row end up living out their lives awaiting execution.}". \texttt{Factivity}, on the other hand, is negatively correlated with almost all the other categories, except \texttt{world}, which means P-H pairs labeled with \texttt{factivity} typically have no other categories marked. 

\section{(Re)Evaluation of SOTA Models}
\label{sec:eval}

 We re-evaluate two Transformer-based and two standard baseline machine learning models on \textsc{TaxiNLI}, under the lens of the taxonomic categories. 
As baselines, we choose BERT-base \cite{devlin2019bert}, and RoBERTa-large \cite{liu2019roberta} as two state-of-the-art NLI systems. For our experiments, we use the pre-trained {BERT-base} and RoBERTa models from HuggingFace's Transformers implementation \cite{Wolf2019HuggingFacesTS}. As pre-Transformer baselines, we use the bidirectional LSTM-based Enhanced Sequential Inference model (ESIM) \cite{chen-etal-2017-enhanced}. We also train a Naive Bayes (NB) model using bag-of-words features for the P-H pairs after removing stop words\footnote{Using NLTK's \texttt{RTEFeatureExtractor}}.

\subsection{\textsc{TaxiNLI} Error Analysis}
We report the NLI task accuracy of the baseline systems on the MNLI validations sets in Table \ref{tab:nliacc}.
The systems are fine-tuned on the MNLI training set using the procedures followed in \citet{devlin2019bert,liu2019roberta,chen-etal-2017-enhanced}. 
\begin{table}[!htpb]
\resizebox{\columnwidth}{!}{
\begin{tabular}{rllll}
\toprule
      \textbf{MNLI-dev} & \textbf{NB} & \textbf{ESIM} & {\bf BERT$_\textsc{base}$} & \textbf{RoBERTa$_\textsc{large}$} \\ \midrule
{Matched} &  51.46  &     72.3   &  84.7         &   92.3      \\ 
{Mismatched} & 52.31 &   72.1   &   84.8        &   90.0      \\ \bottomrule
\end{tabular}
}
\caption{MNLI-validation set accuracy.}
\label{tab:nliacc}
\end{table}

We evaluate the systems on a total of 7.7k examples, which are in the intersection of \textsc{TaxiNLI} and the validation sets of MNLI. 

\begin{figure}[htb!]
	\centering
    \includegraphics[width=0.48\textwidth]{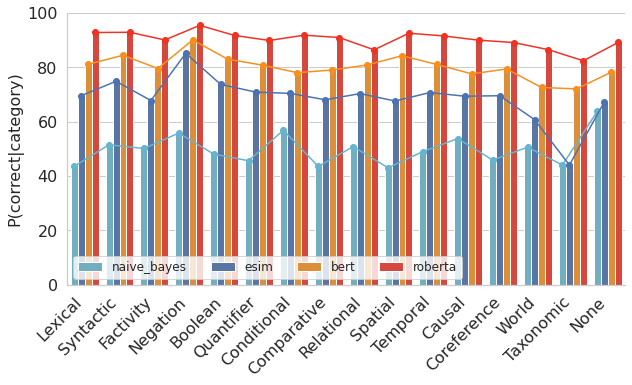}
    \caption{
        The normalized frequency of the systems predicting an example correctly, provided a category. 
    }
    \label{acc}
\end{figure}

Figure \ref{acc} shows
for each category $c_i$, the normalized frequency for a model predicting an NLI example of that category accurately, i.e., $\frac{\#c_i=1,correct=1}{\#c_i=1}$. 
We observe that compared to NB, the improvements in BERT have been higher in \texttt{lexical, syntactic} categories compared to others. Improvements in ESIM compared to NB show a very similar trend, and show for knowledge categories the improvements are negligible. ESIM shows largest improvement on \texttt{negation}.

\subsection{Factor Analysis}
\label{sec:confound}
In order to quantify the precise influence of the category labels on the prediction of the NLI models, we probe into indicators and confounding factors using two methods: linear discriminant analysis (LDA) and logistic regression (LR).
We use indicators for each category (0 or 1) and for two potential confounding variables (lengths of P,{H}), to model the correctness of prediction of the NLI system.
The coefficients of these analyses on BERT are shown in Fig~\ref{fig:factor}.
The values for RoBERTa follow a similar trend, and are presented in the appendix. 
We see that presence of certain taxonomic categories strongly influence the correctness of prediction. 
As we found in the analysis presented in Sec.~\ref{sec:eval},
we observe that {\tt syntactic}, {\tt negation}, and {\tt spatial} categories are strong indicators of correctness of prediction. 
On the other hand, {\tt conditional}, {\tt relational}, {\tt causal}, {\tt coreference} are harder to predict accurately.
Sentence length does not play a significant role.

We also make an observation for categories such as {\tt lexical, syntactic}, where the proportion of a single NLI label is high, also correlated with a high prediction accuracy (Fig.~\ref{fig:factor}).



\begin{figure}
	\centering
    \includegraphics[width=.48\textwidth]{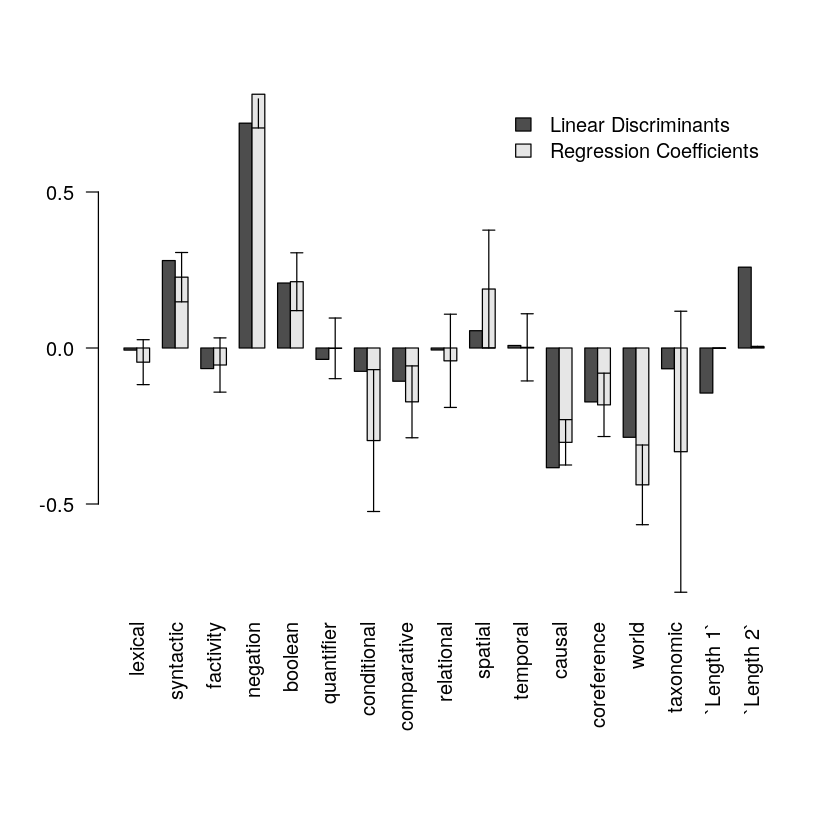}
    \caption{
        Coefficients obtained through Linear Discriminant Analysis (LDA) and Logistic Regression (LR) to model the correctness of NLI prediction by BERT, given taxonomy categories and possible confound variables.
        Significant LR coefficients: 
        {\tt syntactic**}, {\tt negation***, boolean*}, {\tt causal***, world***}, {\tt Length2**}; 
        where $p$ value is smaller than: 0.001***, 0.01**, 0.05*.
    }
    \label{fig:factor}
\end{figure}

\section{Discussion}
\label{sec:discussion}

\definecolor{neutral}{rgb}{1.0, 0.0, 1.0}
\definecolor{entailment}{rgb}{0.75, 0.75, 0.75}
\definecolor{contradiction}{rgb}{0.96, 0.87, 0.7}

\newcommand{\rulesep}{}

\begin{figure*}[htb!]
	\centering
	\begin{subfigure}[t]{0.47\columnwidth}
        \centering
        \includegraphics[width=\textwidth]{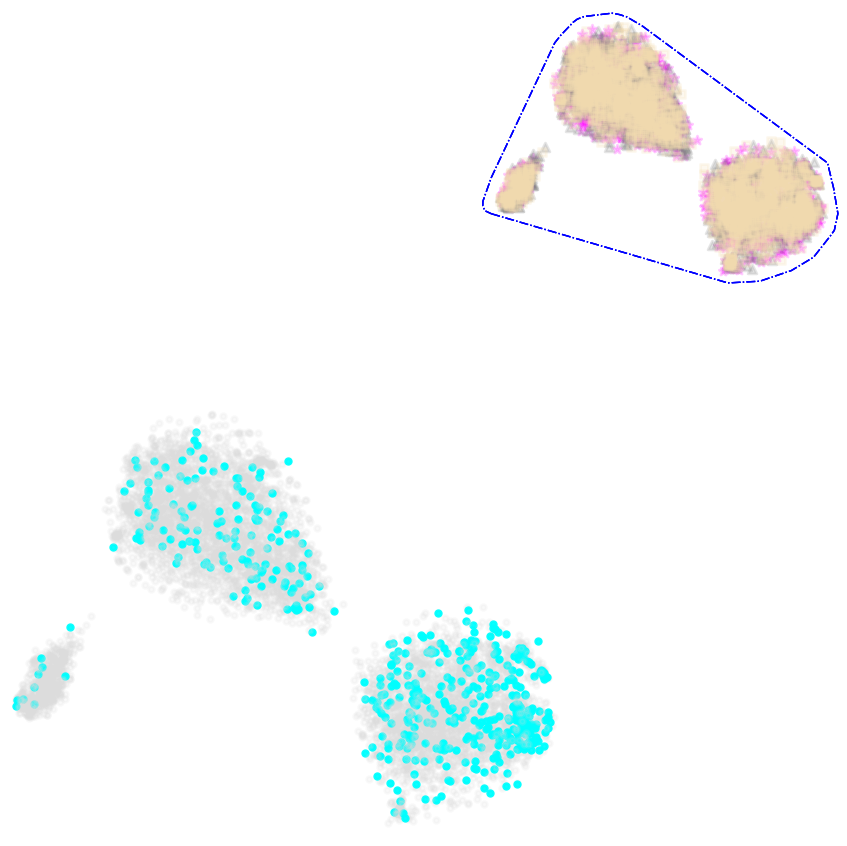}
        \caption{{\tt Lexical} coloured at BERT$_{\rm BASE}$ layer 3}
    \end{subfigure}%
    ~ \rulesep ~
    \begin{subfigure}[t]{0.47\columnwidth}
        \centering
        \includegraphics[width=\textwidth]{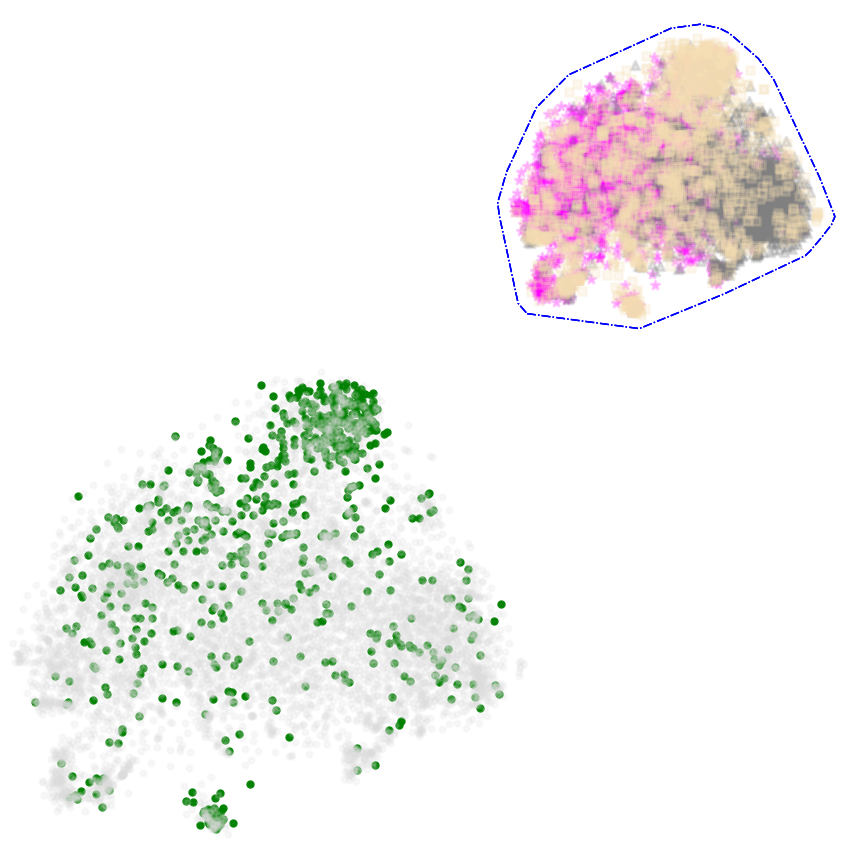}
        \caption{{\it Connectives} coloured at BERT$_{\rm BASE}$ layer 6}
    \end{subfigure}
    ~ \rulesep ~
    \begin{subfigure}[t]{0.47\columnwidth}
        \centering
        \includegraphics[width=\textwidth]{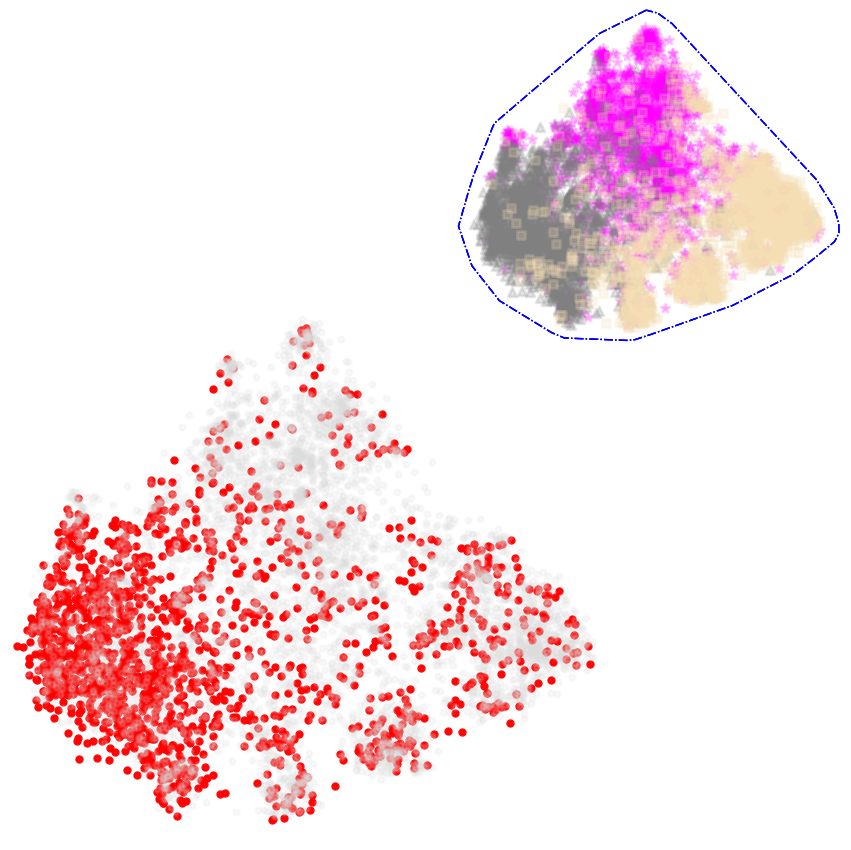}
        \caption{{\tt Syntactic} coloured at BERT$_{\rm BASE}$  layer 11}
    \end{subfigure}
    ~ \rulesep ~
    \begin{subfigure}[t]{0.47\columnwidth}
        \centering
        \includegraphics[width=\textwidth]{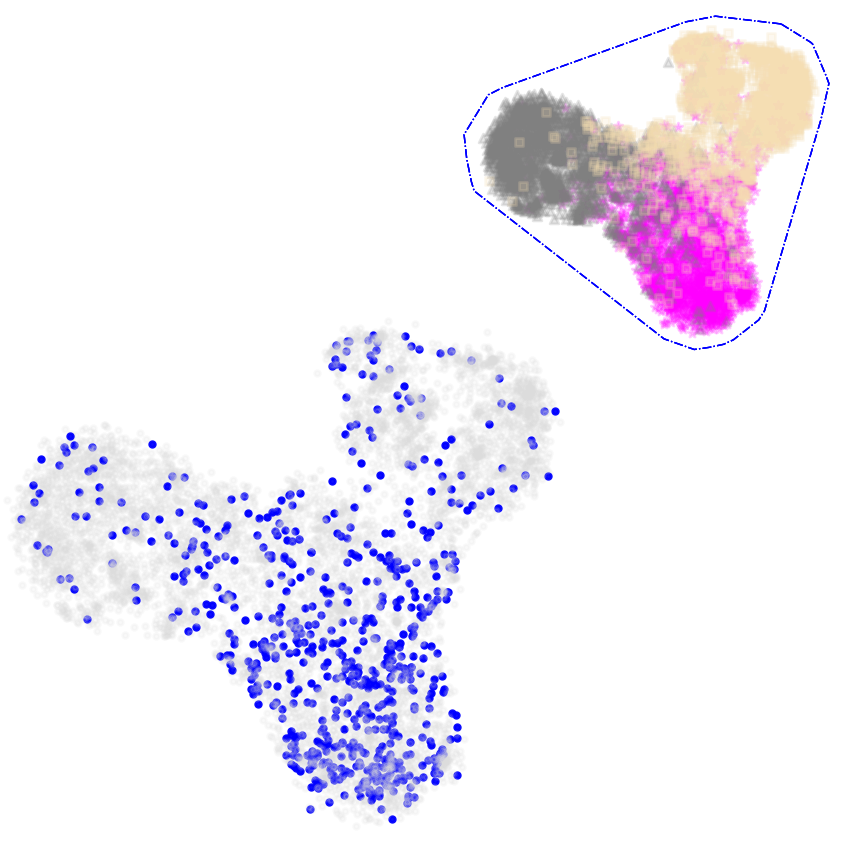}
        \caption{{\it Deductions} coloured at BERT$_{\rm BASE}$ layer 12}
    \end{subfigure}
    %
    %
    %
    %
    %
    \begin{subfigure}[t]{0.47\columnwidth}
        \centering
        \includegraphics[width=\textwidth]{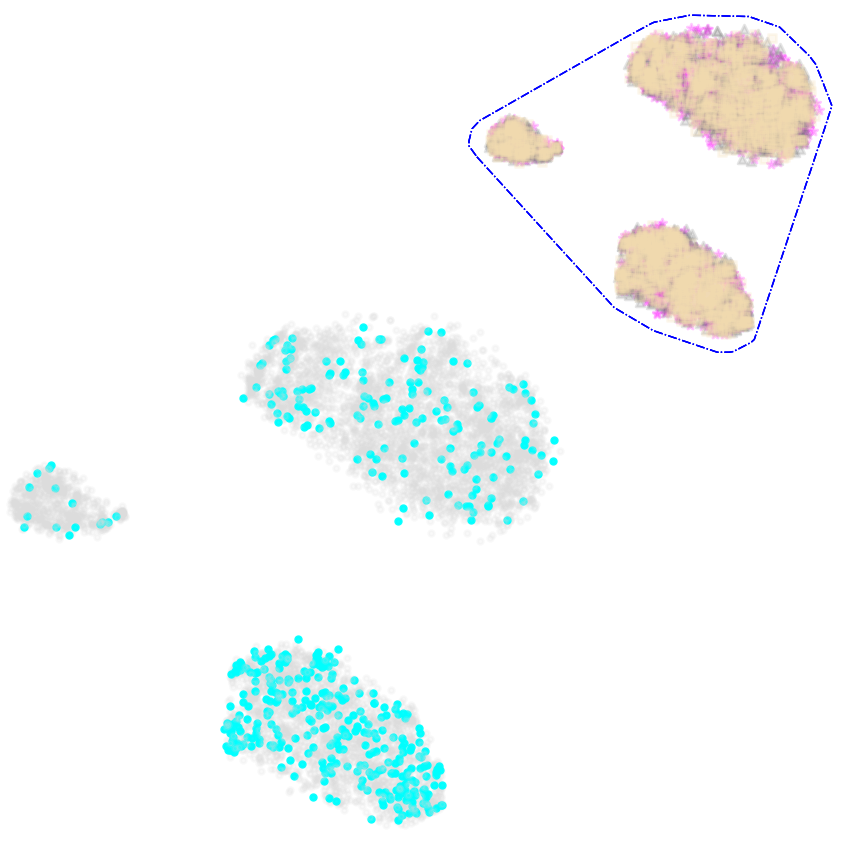}
        \caption{{\tt Lexical} coloured at RoBERTa$_{\rm LARGE}$ layer 1}
    \end{subfigure}%
    ~ \rulesep ~
    \begin{subfigure}[t]{0.47\columnwidth}
        \centering
        \includegraphics[width=\textwidth]{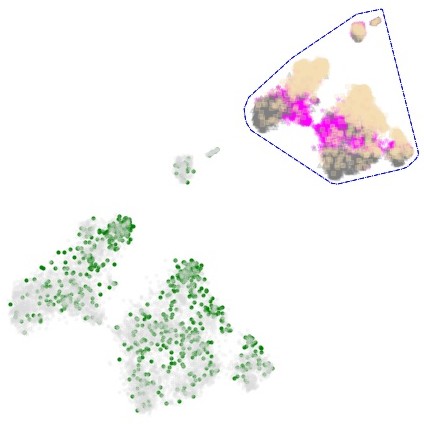}
        \caption{{\it Connectives} coloured at RoBERTa$_{\rm LARGE}$ layer 19}
    \end{subfigure}
    ~ \rulesep ~
    \begin{subfigure}[t]{0.47\columnwidth}
        \centering
        \includegraphics[width=\textwidth]{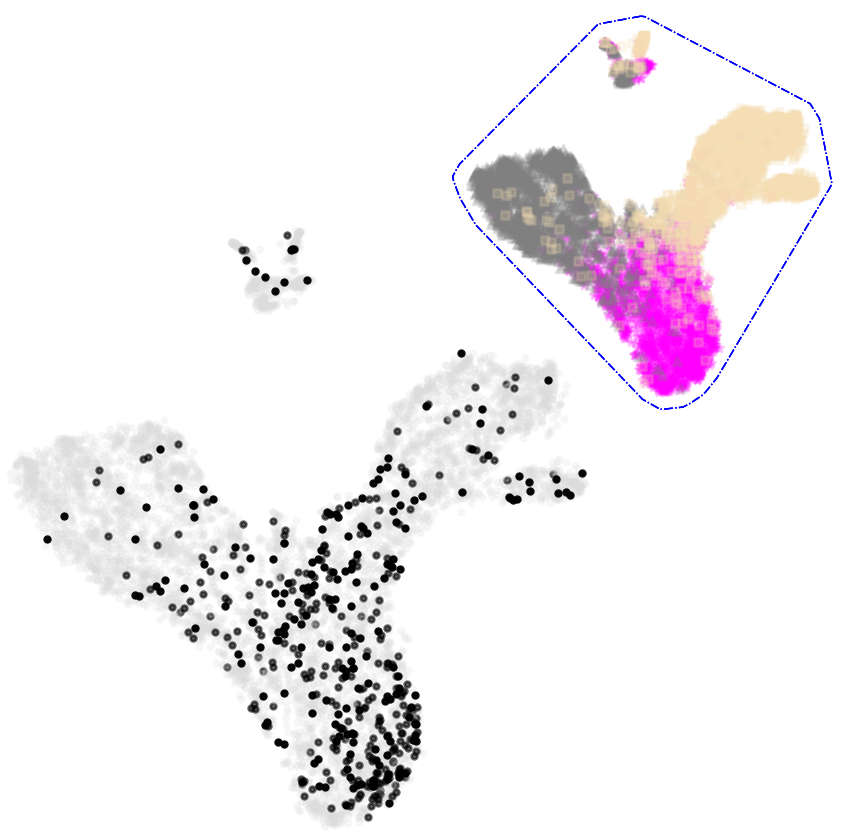}
        \caption{{\it Knowledge} coloured at RoBERTa$_{\rm LARGE}$  layer 21}
    \end{subfigure}
    ~ \rulesep ~
    \begin{subfigure}[t]{0.47\columnwidth}
        \centering
        \includegraphics[width=\textwidth]{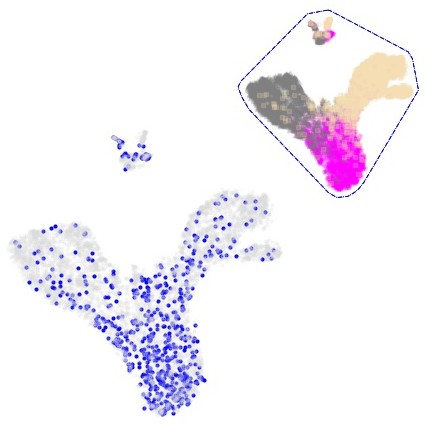}
        \caption{{\it Deductions} coloured at RoBERTa$_{\rm LARGE}$ layer 21}
    \end{subfigure}
    \caption{Layer-wise 2D t-SNE plots of pooled contextualized embeddings of the TaxiNLI examples extracted from BERT and RoBERTA finetuned on MNLI with no retraining on the taxonomic labels. Color codes represent taxonomic categories (inset shows NLI categories: {\textbf{entailment}} (\mycbox{entailment}), \textbf{neutral} (\mycbox{neutral}), \textbf{contradiction} (\mycbox{contradiction})) and their combinations. Only combinations of up to two categories are included for brevity.
    }
    \label{fig:tsne}
\end{figure*}



\paragraph{Visual Analysis}
Section~\ref{sec:eval} paints a thorough picture by analysing the fine-grained capabilities of SOTA NLI systems at a \textit{behavioral}\footnote{
Similar to social sciences, as a black-box system} 
level. Whereas we can say the systems are lacking in certain aspects despite their high overall performance, it naturally also raises questions at the {\it understanding} level:
1) Is there any implicit knowledge acquired by the NLI-finetuned systems about the \textit{kinds} of reasoning required in the inference task? 2) If not, do the systems simply lack the understanding of what kind of reasoning is required per example, or despite understanding that, are unable to do the reasoning?
3) Can we make an argument for future work and model architecture that can more consciously use this information?


In light of recent probing task literature \citep{tenney2019bert,jawahar-etal-2019-bert,liu2019linguistic}, we specifically investigate whether representations of examples cluster meaningfully into taxonomic categories relevant to the reasoning required for NLI.
We use the t-SNE \cite{maaten2008visualizing} algorithm to visualize pooled contextualized representations of NLI examples, under the lens of our taxonomy. For an NLI example, we construct the embeddings at a Transformer layer by
max-pooling hidden states over all input token positions concatenated with the {\tt [CLS]} token (typically used for classification tasks) representation.
The resulting visualizations (Fig.~\ref{fig:tsne}) reveal definitive patterns of clustering by taxonomic categories. The earliest separation is observed for the {\tt lexical} category, at layer 3 in BERT (and layer 1 in RoBERTa), much before any other categories are realized.
At layers 6 in BERT, and 19 in RoBERTa, about the same time as clustering by NLI label is seen, the \textit{connectives} cluster is revealed. The \textit{deductions} (see Sec.~\ref{sec:taxonomy}), and {\tt syntactic} categories are revealed in later layers (layer 11 and 12 in BERT and layer 21 in RoBERTa). The {\it knowledge} category is revealed more prominently in RoBERTa at layer 21, while BERT does not seem to show such a cluster.
By the last few layers, separation into most categories becomes apparent. This means, along various layers of a NLI finetuned language model, taxonomic information is implicitly captured. Despite this, as discussed in the previous sections, SOTA models seem to be deficient in some of the categories---certain categories remain harder to perform inference on.
In the latter layers, the separation along taxonomic categories also corresponds strongly with separation along NLI labels. For instance, in Fig.~\ref{fig:tsne} (c), the examples categorized as {\tt syntactic} almost entirely lie in the {\tt entailment} cloud, which matches our intuition based on the statstics in Fig.~\ref{fig:categories_freq}.

The layer-wise separation of examples by taxonomy raises an interesting possibility to motivate model architectures that may attempt to use its discriminative power to identify such taxonomic categories, for specialized treatment to examples requiring certain reasoning capabilities.




\textbf{Recasting: }
The under-representation of certain categories in the MNLI dataset raises a need for more balanced data collection. 
A possible alternative is to build recast diagnostic datasets for each category, and create probing tasks. Some datasets \cite{zhang-etal-2019-paws,probinglogical} can be recast to the \texttt{syntactic} and \textbf{\texttt{Logical}} categories respectively, as their data creation aligns with our category definitions. However, most categories lack such aligned synthetic data, and crowdsourced data would require manual annotation as above. 
This poses an avenue for future work. 

\section{Conclusion}

To bridge the gap between accuracy-led performance measurement and linguistic analysis of state-of-the-art NLI systems, we propose a taxonomic categorization of necessary inferencing capabilities for the NLI task, and a re-evaluation framework of systems on a re-annotated NLI dataset using this categorization; which underscores the reasoning categories that current systems struggle with.

We would like to emphasize that unlike the case with challenge and adversarial datasets, \textsc{TaxiNLI} re-annotates samples from existing NLI datasets which the SOTA models have been exposed to. Therefore, a lower accuracy in certain taxonomic categories in this case cannot be simply explained away by the ``lack of data" and ``unnatural distribution" arguments.

\section*{Acknowledgements}
We gratefully acknowledge Sandipan Dandapat and Rohit Nargunde for their help regarding annotations. We would like to thank the anonymous reviewers for their insightful comments.

\bibliographystyle{acl_natbib}
\bibliography{emnlp2020}
\appendix

\section{Other Categorizations}
\begin{table}[!htpb]
\resizebox{\columnwidth}{!}{ 
\begin{tabular}{|l|l|l|}
\hline
                                                & ANLI Inference Types                                           & GLUE Diagnostic                                                                                                                                                           \\ \hline
Lexical                                         & \begin{tabular}[c]{@{}l@{}}Standard Inference\\ Lexical Inference\end{tabular}                           & \begin{tabular}[c]{@{}l@{}}Lexical Entailment\\ Morphological Negation\\ Factivity, Redundancy\end{tabular}                                                               \\ \hline
Syntactic                                       & Tricky                                                                                                   & \begin{tabular}[c]{@{}l@{}}Syntactic Ambiguity, \\ Prepositional Phrase \\ Alternations: Active/Passive, \\ Genitives/Partitives, \\ Nominalization, Datives\end{tabular} \\ \hline
\multicolumn{1}{|c|}{\multirow{4}{*}{Semantic}} & Standard Inference                                                                                       & \begin{tabular}[c]{@{}l@{}}Propositional Structure, \\ Intersectivity\\ Quantifiers, Restrictivity, \\ Quantification\end{tabular}                                        \\ \cline{2-3} 
\multicolumn{1}{|c|}{}                          & \begin{tabular}[c]{@{}l@{}}Numerical, \\ Quantitative\end{tabular}                                       & Counting                                                                                                                                                                  \\ \cline{2-3} 
\multicolumn{1}{|c|}{}                          & Reference and Names                                                                                      & \begin{tabular}[c]{@{}l@{}}Symmetry/Collectivity, \\ Coreference, Richer \\ Logical Structures\end{tabular}                                        \\ \cline{2-3} 
\multicolumn{1}{|c|}{}                          & \multicolumn{1}{c|}{\begin{tabular}[c]{@{}c@{}}Reference and Names\\ Reasoning about Facts\end{tabular}} & \begin{tabular}[c]{@{}l@{}}Named Entities \\ Knowledge and \\ Commonsense\end{tabular}                                                                                    \\ \hline
Pragmatics                                      & Tricky                                                                                                   & Ellipsis/Implicits                                                           \\ \hline
\end{tabular}
}
\caption{We show existing NLI error-analysis categorizations proposed by the recent papers, and group them in higher-level categories.}
\label{tab:others}
\end{table}

\section{Bayesian Estimate of Correctness Correlation with Categories}
Since, examples are annotated with multiple categories, we capture the dependencies by defining a Bayesian Network (BN) where, each category (a boolean random variable) has a directed edge to the \textit{correct} node (representing correctness of prediction)\footnote{Additionally, we attempted to learn a Bayes Net from the data using \texttt{bnlearn} package. But the limited number of observations yield non-intuitive results.}. We learn the parameters by fitting this BN to the observed data. In Figure \ref{acc2}, we see from the Bayesian estimate again that, the improvements by BERT in categories such as \texttt{relational} reasoning has been low. It also shows, that there is a sharp decrease in accuracy for examples requiring the use of \texttt{taxonomic} knowledge. However, RoBERTa improves over NB and ESIM by large margins, albeit non-uniformly. 

\begin{figure}[htb!]
	\centering
    \includegraphics[width=0.48\textwidth]{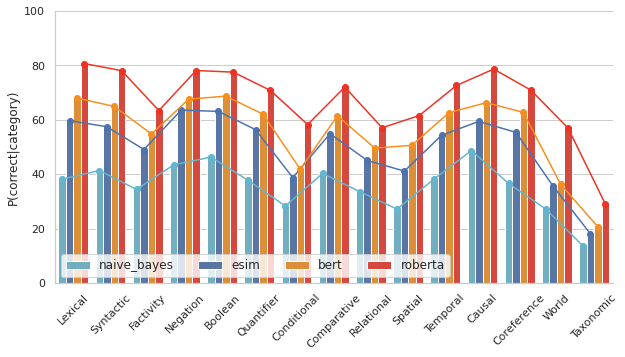}
    \caption{We show a Bayesian Estimate of $P({\tt correct}=1|{\tt category}=1)$ for different systems. }
    \label{acc2}
\end{figure}

\section{Factor Analysis of Correctness of Prediction by RoBERTa}

\begin{figure}
	\centering
    \includegraphics[width=0.48\textwidth]{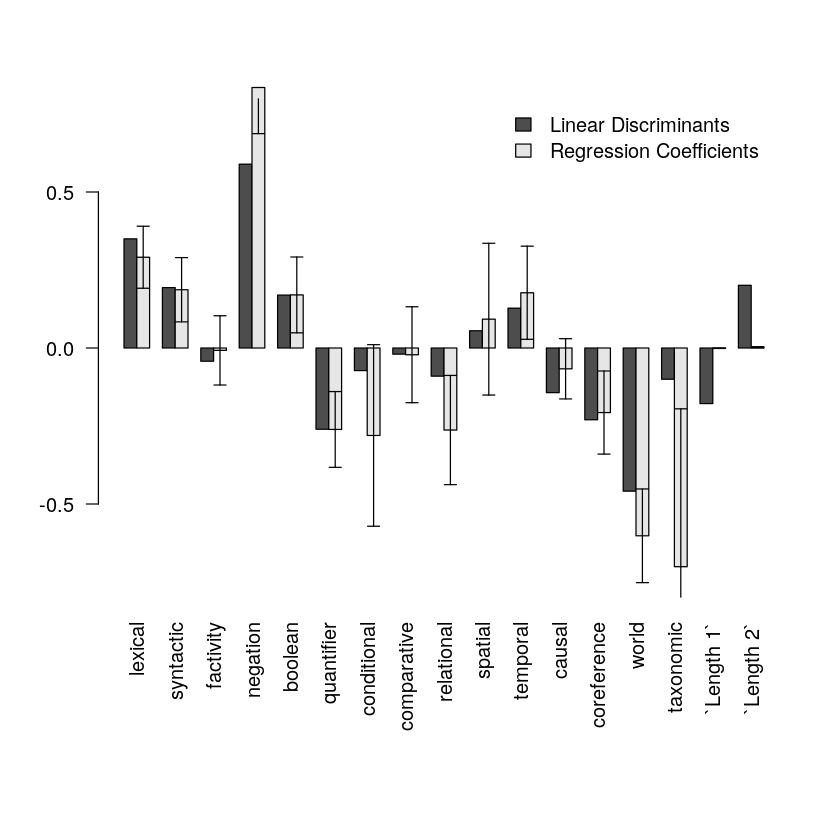}
    \caption{Coefficients obtained through Linear Discriminant Analysis (LDA) and Logistic Regression (LR) to model the correctness of NLI prediction by RoBERTa, given taxonomy categories and possible confound variables.
        Significant LR coefficients: 
        {\tt lexical**}, {\tt negation***, world***}, {\tt quantifier*, boolean*}, {\tt Length1*}; 
        where $p$ value is smaller than: 0.001***, 0.01**, 0.05*.}
    \label{acc2}
\end{figure}
In Fig.~\ref{acc2}, we show the results of Linear Discriminant Analysis (LDA) and Linear Regression (LR) results for RoBERTa predictions. A very similar trend as BERT can be seen here as well.

\section{Annotation Questionnaire}
Our annotation process went through several steps of refinement and improvement. We started with the most basic annotation flow, which was to have a manual which defines each taxonomic category in detail, and then have the annotator mark for each category. For the pilot study, we took roughly 300 examples from MNLI and asked an initial annotator to annotate. The feedback was the following:
\begin{itemize}
    \item The manual describing each taxonomic category had a lot of information and took time to understand and digest.
    \item It was difficult to keep referring to the guide, although after sufficient examples, it became easier.
    \item There was confusion and ambiguity about the definitions, and the annotator interpreted the definitions differently than what we intended.
    \item Figuring out the categorical annotations for neutral examples was a challenge, as sometimes the topic or subject of what the hypothesis was discussing was separate from what the premise was discussing. 
\end{itemize}
Through the analysis of these annotations, we also observed that some of the initial categories we had were either exceedingly underrepresented in the MNLI dataset, or were consistently confused with others. Thus, we revised the set of categories, setting more distinct boundaries, and ensuring independence of categories. We revised the questionnaire into a hierarchical 'if-else' multi-choice design. The questionnaire is structured as follows:

\subsubsection{Questionnaire 1}
``We present you with a set of statements. Statement 1 (S1) is the truth and context. Statement 2 (S2) is a claim/hypothesis. The task is to evaluate statement 2 as true, false, can't say.

S1: ...
S2: ...

\begin{enumerate}
    \item Can you evaluate S2 by just using the information/context given in S1? Or do you require knowledge from external documents, say history books, news articles, science books, etc.?
    \begin{enumerate}
        \item Need more information
        \item Do not need more information
    \end{enumerate}
    
    \item If yes, what kind of information did you require? (More than one answer can be ticked)
    \begin{enumerate}
        \item Knowledge about certain facts from say history books, news articles, tech magazines, etc.? This is also knowledge about named entities (e.g. Obama, Taj Mahal, New York etc.). E.g:
        \begin{itemize}
            \item S1: Barack Obama lived in the White House during 2009-17.
            \item S2: Barack Obama was the President in 2009. 
        \end{itemize}
        This is TRUE and requires external knowledge that US presidents live in the White House.
        
        \item Knowledge about taxonomies and hierarchies. A few examples are animal groups (snakes are reptiles), currencies (dollar is a currency), types of activities (football is a sport, sport is an activity).  Basically, how a common noun (snake) belongs to a class (say reptiles), which can belong to yet another class (animals). Do not select this if the name of one object belongs (or is a substring) of the other class of objects (e.g. green snake is a snake isn't part of this category), or if the names of the objects are pronouns (e.g. Barack Obama - president and related examples are part of category 2a, not this one). E.g:
        \begin{itemize}
            \item S1: Norman hated all musical instruments.
            \item S2: Norman loves the piano.
        \end{itemize}
        This is FALSE and requires external knowledge that a piano belongs to the class of instruments, hence Norman cannot love the piano.
        
        \item No extra knowledge required
    \end{enumerate}
    
     \item Using just the information from S1, and given that you have the required knowledge from the above question, did you have to use some reasoning to figure out the answer, or did you just need the knowledge of words and paraphrasing, or both? (More than one can be ticked)
    \begin{enumerate}
        \item Some reasoning was required, which wasn't explicitly written down in S1, but was implicitly understood.
        
        \item Knowledge of words (e.g. synonyms, antonyms), and recognizing paraphrases. The information I needed was explicitly written down in S1.
    \end{enumerate}
    
    \item What kind of reasoning was required (if applicable) (More than one can be ticked)?
    \begin{enumerate}
        \item You needed reasoning about relations in S1. You observed that there were objects/entities in S1 and there were explicit mentions of how they were related (e.g. Jack and his son went to the circus), and you used your reasoning about the nature of those relations to arrive to the answer (e.g. Jack and a stranger went to the circus is FALSE) E.g:
        \begin{itemize}
            \item S1: Jack and his son went to the circus. 
            \item S2: Jack and a stranger went to the circus. 
        \end{itemize}
        This is FALSE, but you need to reason that S1 contains the relation “father of” between Jack and some person X (who is his son). S2 is false because if X is a stranger, it cannot be Jack’s son.
        
        \item You needed reasoning about spatial setup. S1 contained information about relative locations of objects/entities (e.g. Jack was on the right of Jim, and Jim was on the right of John) and you needed to reason about how objects/entities were located, when it wasn't explicitly stated (e.g. John was on the left of Jack is TRUE).
        
        \item You needed reasoning about time intervals, duration, or temporal reasoning. S1 contained information about events and time information (e.g. Jack went to the shop from 8:00am to 10:00am), and you needed to reason about the timing to arrive at the answer (e.g. Jack was at the shop at 9:30am is TRUE). 
        
        \item You needed reasoning about cause, effect, and intent behind it. S1 contained information about an event or an action (e.g. Jack was hurt). S2 contains either a cause or an intent, and you needed to reason whether S1 and S2 were possible cause/effect or intent/effect pairs (e.g. Jack was hit by a car is CAN'T SAY)
        \begin{itemize}
            \item S1: X shot Y
            \item S2: Y is hurt
        \end{itemize}
        This is TRUE, but you needed to reason that upon getting shot, Y should get hurt. 
        
        \item You needed to be able to reason about who is being referred to in the text. Better demonstrated via example: 
        \begin{itemize}
            \item S1: Jane didn't visit Janette because \underline{she} didn't want to speak with her. 
            \item S2: Jane didn't want to speak with Janette.
        \end{itemize}
        This is TRUE. ‘She’ refers to Jane, and you needed to reason about that to get the answer.
        
    \end{enumerate}
    
    \item Did you need logical reasoning? This applies if S1 and/or S2 consist of statements connected by logical connective words (and, or, not, every, some, only, either, neither, etc. or any synonyms of these words). These connectives were important for arriving at the answer. If so, which connectives? (Can tick multiple choices)
    \begin{enumerate}
        \item Negation (not, no, \underline{in}capable etc.), where S2 negates one of the facts in S1. E.g:
        \begin{itemize}
            \item S1: Laurie has visited Nephi, Marion has only visited Calistoga.
            \item S2: Laurie didn’t visit Nephi.
        \end{itemize}
        This is FALSE, and S2 is a negation of the first statement in S1.
        
        \item Boolean (or, and), where S1 is a set of statements connected by Or and AND, and S2 talks about one or more of these statements. E.g:
         \begin{itemize}
            \item S1:  Jar Jar Binks, R2D2 and Padme only visited Anakin’s house.
            \item S2: Jar Jar Binks didn’t visit Anakin’s shop. 
        \end{itemize}
        This is TRUE, and S1 is connected by ‘and’ statements for three entities who visited Anakin’s house. S2 talks about the sub-statement “Jar Jar Binks only visited Anakin’s house” in the ‘and’ connective, and it is true because Jar Jar Binks didn’t visit anywhere else. 
        
        \item Quantifier (every, some, at least, at most, etc.), where S1 and S2 contain the use of these terms.
        \begin{itemize}
            \item S1: Everyone visited Anakin’s home.
            \item S2: Padme didn’t visit Anakin’s home.
        \end{itemize}
        This is FALSE, with S1 containing the quantifier ‘everyone’, and S2 stating that someone, Padme, didn’t visit.
        
        \item Conditionals (if-else, if-then, etc.), where S1 has if-then, if-else statements or similar.
        \begin{itemize}
            \item S1:  Francisco has visited Potsdam and if Francisco has visited Potsdam then Tyrone has visited Pampa.
            \item S2: Tyrone has visited Pampa
        \end{itemize}
        This is TRUE, since there is a if-then condition in S1, and it is satisfied to make S2 true. 
        
        \item Comparatives (e.g. as tall as, taller than, faster than, etc.) where S1 compares entities via these comparative phrases, and S2 needs knowledge about the comparisons.
        \begin{itemize}
            \item S1: John is taller than Gordon and Erik, and Mitchell is as tall as John
            \item S2: Gordon is taller than Mitchell.
        \end{itemize}
        This is FALSE. This are comparative statements “Is taller than” in S1, and S2 needs logical reasoning on who is taller than whom to get S2. In addition to this, this also needs knowledge of Boolean due to the presence of the ‘and’ in S1.
        
    \end{enumerate}
    
    \item Finally, can you describe some word/phrase (explicitly written) properties of S1 and S2 which helped to arrive to the answer? (more than one can be ticked)
    \begin{enumerate}
        \item S1 and S2 were almost the same, apart from the removal, addition, or substitution of a few words. If substituted, the words were synonyms or antonyms. E.g:
        \begin{itemize}
            \item S1: Anakin Skywalker was compassionate.
            \item S2: Anakin Skywalker was cruel.
        \end{itemize}
        This is FALSE, with S1 and S2 being very similarly framed statements, with the substitution of a word for it’s antonym. Thus it belongs to this category.
        
        \item S1 and S2 were paraphrases of each other. S2 is a paraphrase of S1 or a certain part of information mentioned in S1
         \begin{itemize}
            \item S1: Anakin was an excellent pilot.
            \item S2: The piloting skills of Anakin were excellent.   
        \end{itemize}
        This is TRUE, and S1 and S2 being paraphrases of one another. Also, to note, if ‘excellent’ in S2 were replaced by ‘terrible’, it would still fit this category, but would also fit category 6a, since it would be a paraphrase with a swapped word.
        
        \item S2 contains an assumed fact from S1, mostly an assumption about the existence or the occurrence of an action. 
        \begin{itemize}
            \item S1: Anakin found the Death Star.
            \item S2: The Death Star exists.
        \end{itemize}
        This is TRUE. The Death Star exists if Anakin has found it, thus S1 makes the assumption that it exists.
        \begin{itemize}
            \item S1: James was happy that his plane could fly.
            \item S2: His plane couldn’t fly.
        \end{itemize}
        This is FALSE. Since James was happy that the plane flew (S1 makes the assumption that it happened), it is FALSE that his plane couldn't fly since it happened.''
        
    \end{enumerate}
\end{enumerate}

The above questionnaire was given along with premise-hypothesis pairs having the gold label of `entailment' or `contradiction' . However, to prevent biasing the annotator, we allowed them to choose `neutral' (CAN'T SAY) as well. 

\subsubsection{Questionnaire 2}
This questionnaire was given to the annotators after they had done a sufficient number of `entailment/contradiction' samples using Questionnaire 1. For Questionnaire 2, annotators were told that the datapoints were `neutral', and asked them to first answer these 3 questions:

``Given S1, there isn’t enough information to decide whether S2 is TRUE or FALSE. Please answer the following questions for each datapoint which has been annotated as CAN’T SAY.

\begin{enumerate}
    \item Are S1 and S2 talking about the same general topic (e.g. sports, politics, religion)? [Yes/No]
    \item If yes, are S1 and S2 talking about the same subject? (e.g. S1: Obama was the president of USA, S2: Obama was a nice guy, subject of sentence is Obama) [Yes/No]
    \item If yes, are S1 and S2 talking about the same objects of discussion? (e.g. S1: Obama lived in the White House often, S2: Obama said the White House was huge. Here both subject (Obama) and object (White House) of discussion are the same)
    \item If yes, what kinds of information are required which you used, and what kinds of information are missing? If no, what kinds of information are required which you used, and what kinds of information are missing? ''
\end{enumerate}

Upon answering the above, if the answer to the second question was yes, then they proceeded with the category annotation, else they moved on to the next question. This helped eliminate the random hypotheses.

\subsubsection{Annotator Feedback}
We received a lot of important feedback from our annotators during the clarification and training sessions. They are listed below:
\begin{itemize}
    \item  Many premise sentences seems out of place, and the context is still insufficient many times. As a result, the hypothesis also introduces ambiguity, making the process a bit tricky.
    \item There were cases where a certain name of an entity in the premise is switched for something else in the hypothesis. This created some confusion because it fell somewhere between lexical and coreference (according to the annotator). 
    \item Another confusion arose from the quantifier category, where initially the name of the category led the annotators to believe that it referred to not just what we described (e.g. some, all), but quantities (say 5000 in the premise was swapped with 2000 in the hypothesis). This was again a middle ground between lexical and quantifier.
    \item Many of the premises contained incoherent, difficult to understand sentences. A lot of premises (which we later found to be from the telephone category), contained many filler words (uh, uhm, etc.) which made comprehension difficult.
    \item Another issue lies with an implicit rigidity of the annotation process using just the questionnaire. The targeted questions were written so as to allow annotators to generalize and apply intuitive principles along those thought lines that we try to demarcate via the questions. We wanted to prevent them completely However, as annotators have not been exposed to the exact intentions behind the annotation (so as to prevent bias), they followed the questionnaire strictly, and did not always generalize until subsequent training/clarification sessions where we encouraged them to generalize. However, the implicit rigidity still impacts the annotations to some extent, although mitigated to a large level by the training. This remains a challenge due to the tradeoff between open interpretation of the task, as well as a rigidity of desired annotations which stem from an analysis perspective (from our side). 
    \item Idiomatic references, metaphors, and common phrases were also a source of confusion, and although to some extent were marked as world knowledge, did leave annotators unsure about where to place them.

\end{itemize}
The above feedback only strengthened our belief in an iterative training system for a complicated task such as this. It also sheds light on how difficult a task like this is to crowdsource.




\end{document}